\documentclass[letterpaper, 10 pt, journal, twoside]{ieeetran}                              
\usepackage{cite}
\usepackage{amsmath,amssymb,amsfonts}
\usepackage{algorithmic}
\usepackage{graphicx}
\usepackage{textcomp}
\usepackage{bm}
\usepackage{booktabs}
\usepackage{pifont}
\usepackage{graphicx}
\usepackage{caption}
\usepackage{xcolor}
\usepackage{orcidlink}
\usepackage{makecell}
\usepackage{threeparttable}
\usepackage{hyperref} 
\hypersetup{
  colorlinks   = true,    
  urlcolor     = blue,    
  linkcolor    = red,    
  citecolor    = blue      
}

\def\BibTeX{{\rm B\kern-.05em{\sc i\kern-.025em b}\kern-.08em
    T\kern-.1667em\lower.7ex\hbox{E}\kern-.125emX}}

\begin{document}

\title{MFE: A Multimodal Hand Exoskeleton with Interactive Force, Pressure and Thermo-haptic Feedback}

\author{Ziyuan Tang\orcidlink{0009-0004-0509-8307}, Yitian Guo\orcidlink{0009-0009-2126-0008} and Chenxi Xiao\orcidlink{0000-0002-7819-9633}$^{*}$
\thanks{Manuscript received: December, 17, 2025; Accepted January, 22, 2026.}
\thanks{This paper was recommended for publication by Editor Kyung Ki-Uk upon evaluation of the Associate Editor and Reviewers' comments. This work was supported by the Natural Science Foundation of Shanghai (Grant No. 25ZR1402370), and partially by Shanghai Frontiers Science Center of Human-centered Artificial Intelligence (ShangHAI), MoE Key Laboratory of Intelli-gent Perception and Human-Machine Collaboration (KLIP-HuMaCo).}
\thanks{Open-source files of this work are available at \href{https://github.com/TangRobot/MFE}{github.com/TangRobot/MFE}}
\thanks{All authors are with the School of Information Science and Technology, ShanghaiTech University, Shanghai, China (*corresponding author, {\tt\footnotesize tangzy2022, guoyt2022, xiaochx@shanghaitech.edu.cn}).}
\thanks{Digital Object Identifier (DOI): 10.1109/LRA.2026.3662616}
}

\maketitle

\begin{abstract}
Recent advancements in virtual reality and robotic teleoperation have greatly increased the variety of haptic information that must be conveyed to users. While existing haptic devices typically provide unimodal feedback to enhance situational awareness, a gap remains in their ability to deliver rich, multimodal sensory feedback encompassing force, pressure, and thermal sensations. To address this limitation, we present the \textbf{Multimodal Feedback Exoskeleton (MFE)}, a hand exoskeleton designed to deliver hybrid haptic feedback. The MFE features 20 degrees of freedom for capturing hand pose. For force feedback, it employs an active mechanism capable of generating 3.5-8.1~N of pushing and pulling forces, enabling realistic interaction with deformable objects. The fingertips are equipped with flat actuators based on the electro-osmotic principle, providing pressure and vibration stimuli and achieving up to 2.47~kPa of contact pressure to render tactile sensations. For thermal feedback, the MFE integrates thermoelectric heat pumps capable of rendering temperatures from 10\,$^\circ$C to 55\,$^\circ$C. We validated the MFE by integrating it into a robotic teleoperation system using the X-Arm~6 and Inspire Hand manipulator. In user studies, participants successfully recognized and manipulated deformable objects and differentiated remote objects with varying temperatures. These results demonstrate that the MFE enhances situational awareness, as well as the usability and transparency of robotic teleoperation systems.

\end{abstract}


\section{Introduction}
\IEEEPARstart{T} {eleoperation} systems enable the remote control of robots by replicating human movements, leveraging robotic dexterity through the telepresence of user’s control and sensory feedback \cite{ozdamar2022shared}. These systems support diverse tasks in environments that are unsafe or inaccessible to humans. For instance, they are employed in robotic laparoscopic surgery that needs to be accomplished minimally invasive \cite{patel2022haptic}, and in hazardous scenarios where human safety is a primary concern \cite{xiao2023tactile}. More recently, teleoperation has become essential for data collection in embodied AI research \cite{luo2024humanagent}. High-fidelity data capture interfaces, such as exoskeletons, have improved teleoperation performance in both the efficiency and quality of data acquisition \cite{cheng2024tv, wu2024humanft}. Across all these applications, the need for intuitive and immersive control interfaces for robots is becoming increasingly critical.

Haptic devices are essential components of teleoperation interfaces. Haptic feedback transmits contact information from remote environments to user's side, directly contributing to improved situational awareness, safety, and task performance \cite{opiyo2021review, riley2004situation, Wildenbeest2013HapticImpact}. To support this capability, conventional teleoperation platforms have developed various haptic feedback interfaces, such as joysticks and exoskeletons, that renders the contact states from either bi-fingered grippers or dexterous robotic hands \cite{Lach2022Grasping, Ren2023Interventional, Lenz2021Bimanual}. However, most of these designs primarily focus on vibrotactile or force feedback, offering limited modality and bandwidth. This falls short of the multimodal richness of human skin perception, which integrates diverse sensory inputs including force, pressure, temperature, pain, and itch.

\definecolor{Daffodil}{rgb}{0.87, 0.72, 0.53}
\definecolor{forestgreen(web)}{rgb}{0.13, 0.55, 0.13}
\begin{figure}
    \centering
    \includegraphics[width=1\linewidth]{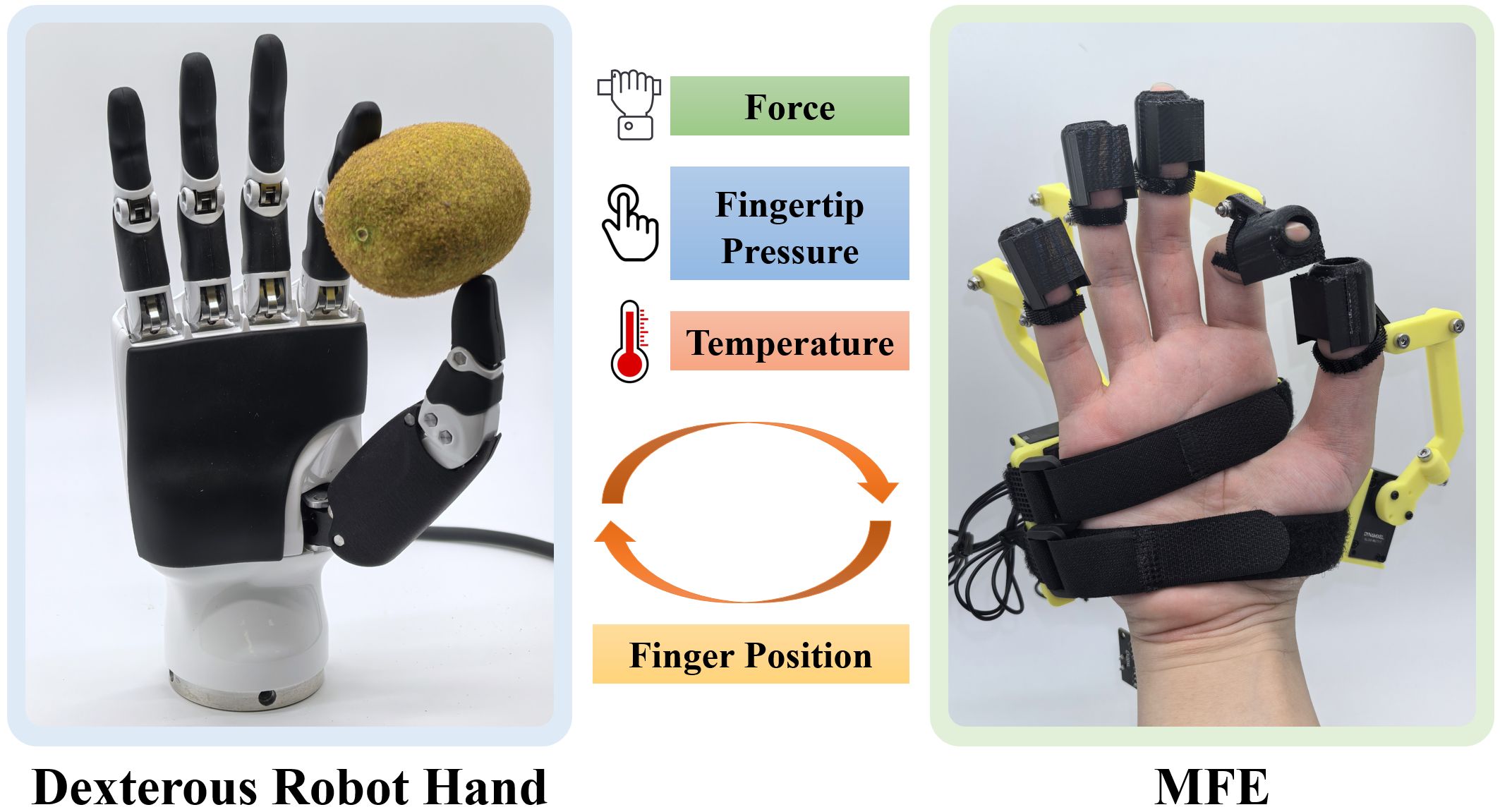}
    \caption{Our exoskeleton is capable of capturing \textcolor{Daffodil}{finger positions} and rendering remote \textcolor{forestgreen(web)}{contact force}, \textcolor{blue}{fingertip pressure}, and \textcolor{orange}{palm temperature}.}
    \label{main_pic}
\end{figure}

Recent advances in embodied AI and virtual reality have also called for haptic devices capable of rendering a broader modalities of touch data. Researchers have collected various forms of tactile data suitable for haptic replication, including object surface textures \cite{CAO2024Multimodal}, thermal and acoustic cues \cite{gao2022objectfolder}, and vibration and pressure sensations during contact and sliding motions \cite{wu2024humanft}. Beyond datasets, recent work has also focused on developing world models capable of generating multiphysics effects \cite{Fu2023multiphysics}. To support multimodal tactile rendering, recent efforts have explored techniques for force-feedback  \cite{Lenz2021Bimanual} and pressure rendering \cite{Pascher2023HaptiX}. Nevertheless, most existing designs typically support only a single haptic modality, are prohibitively expensive, or lack open-source implementations. These limitations underscore the need for a unified and accessible solution for interactively rendering multiphysics effects.

To address these challenges, we introduce the Multimodal Feedback Exoskeleton (MFE): a low-cost, and open-source platform designed to render multimodal haptic effects. MFE integrates a suit of haptic rendering techniques to provide hybrid feedback, encompassing active force feedback to the fingers, pressure feedback at fingertips, and temperature feedback at the palm, making high-fidelity haptic interaction more accessible to researchers and practitioners. This is achieved through several key technical innovations. First, MFE features an active force-feedback mechanism for each finger, capable of generating both pulling and dragging forces. While such functionality was previously limited to high-end exoskeletons (e.g., \cite{senseglove}), our design offers a cost-effective alternative by combining force-controlled motors with ergonomically customizable linkage mechanisms and force-mapping algorithms. Second, the system employs electro-osmotic flat actuators to deliver fingertip pressure and vibration feedback, enabling responsive and realistic contact sensations. Additionally, thermoelectric heat pumps are integrated to provide dynamic temperature feedback, revealing thermal cues of manipulated objects. Finally, we evaluate MFE through comprehensive device characterization and user studies, demonstrating its effectiveness in teleoperation scenarios.

In summary, our key contributions are as follows:
\begin{enumerate}
\item \textbf{MFE:} A low-cost hand exoskeleton capable of delivering hybrid haptic feedback that combines force, pressure, and temperature modalities.

\item Haptic rendering techniques for rendering multimodal haptic effects, including active force-feedback mechanism, microfluidic flat actuators for pressure rendering, and thermoelectric heat pump for temperature rendering.

\item Experimental validation of MFE through device characterization and user studies, and discussions of its usability as teleoperation interfaces.
\end{enumerate}

\section{Related Works}

\subsection{Haptic Rendering Devices} 
Haptic feedback aims transmitting remote contact sensation to human operators via rendering devices. Among these, force feedback is fundamental, as interactions in every simulated and real physics follow Newtonian mechanics \cite{kitagawa2002analysis, rebelo2014bilateral, xiao2021fingers}. However, force feedback alone is insufficient for conveying comprehensive contact information. To compensate this, recent work has focused on rendering skin contact pressure through pneumatic systems \cite{du2024haptiknit, Gerald2022Colonoscopy}, fluid-based mechanisms \cite{shen2023fluid}, and magnetic actuation \cite{Lu2022Magnetic}. Beyond force and pressure, vibrotactile feedback supplements high-frequency details, such as surface roughness and texture \cite{junput2019feel}. Thermal feedback, using liquid or air-based heating and cooling systems, has also been simulated \cite{cai2020thermairglove, han2018hydroring, kim2020thermal}. Despite these advancements, most existing haptic systems still focus on individual modalities, limiting their ability to provide a comprehensive haptic experience. To bridge this gap, we present a system capable of rendering force, pressure, and temperature feedback, contributing to the development of multimodal haptic systems.

\subsection{Interfaces for Teleoperating Dexterous Hands}
Recent advancements in the embodied AI community have focused on teleoperating dexterous robotic hands in both physical environments and virtual reality realms. To control high-degree-of-freedom manipulators with human-like dexterity, two main teleoperation approaches have emerged. The first approach relies on vision-based hand-tracking techniques \cite{Mizera2020Evaluation}, which are easy to deploy due to minimal wearable hardware requirements. However, these systems lack haptic feedback, limiting the operator’s situational awareness during contact-rich tasks \cite{qin2023anyteleop, handa2020dexpilot, li2019vision}. To address this limitation, the second approach incorporates wearable exoskeletons capable of delivering haptic feedback. For example, the iCub humanoid robot has been integrated with the SenseGlove DK1 \cite{senseglove} to transmit finger forces to users \cite{dafarra2024icub3}. Similarly, the Shadow Dexterous Hand, when combined with the HaptX glove \cite{haptx_website}, can provide both force and pressure feedback. However, such systems either lack active force feedback for finger retraction \cite{senseglove} or are bulky and prohibitively expensive \cite{shadowrobot_teleoperation}, limiting their accessibility to the broader research community (Table \ref{compare}). Moreover, existing devices rarely support the transmission of temperature sensations during teleoperation, which are essential for perceiving object's material properties and safety \cite{ZHAO2025111255}. To overcome these limitations, our work aims to develop a low-cost exoskeleton that provides multimodal haptic feedback for enhancing situational awareness and is accessible to practitioners in both robotics and virtual reality.

\begin{table}[h]
\vspace{+1mm}
    \centering
    \caption{Comparison of different  hand's haptic interfaces.} 
    \begin{tabular}{c | c c c}
        \hline\hline
         Interface & Motion Cap & Feedback & Cost \\
        \midrule
        SenseGlove DK1 \cite{senseglove} & 20 DoFs & \makecell{Resistant Force\\Fingertip Vibration} &\$3524 \\
        \midrule
        HaptX Gloves G1 \cite{haptx_website} & 36 DoFs & \makecell{Active Force \\Fingertip Pressure} &\$5495 \\
        \midrule
        LucidGloves \cite{lucidgloves} & 5 DoFs & No & \$60\\
        \midrule
        DOGlove \cite{zhang2025doglove}& 21 DoFs & \makecell{Active Force\\Fingertip Vibration} & \$600\\
        \midrule
        MFE (Ours) & 20 DoFs & \makecell{Active Force\\Fingertip Pressure\\Temperature} & \$350\\
        \hline\hline
    \end{tabular}
    \begin{tablenotes}
      \footnotesize
    \item The cost of DK1 and HaptX Gloves G1 represents the market price, while the cost of the remaining items correspond to material cost. 
    \end{tablenotes}
    \label{compare}
\end{table}
 
\section{Methodology}

\begin{figure*}[t]
    \centering
    \includegraphics[width=0.98\linewidth]{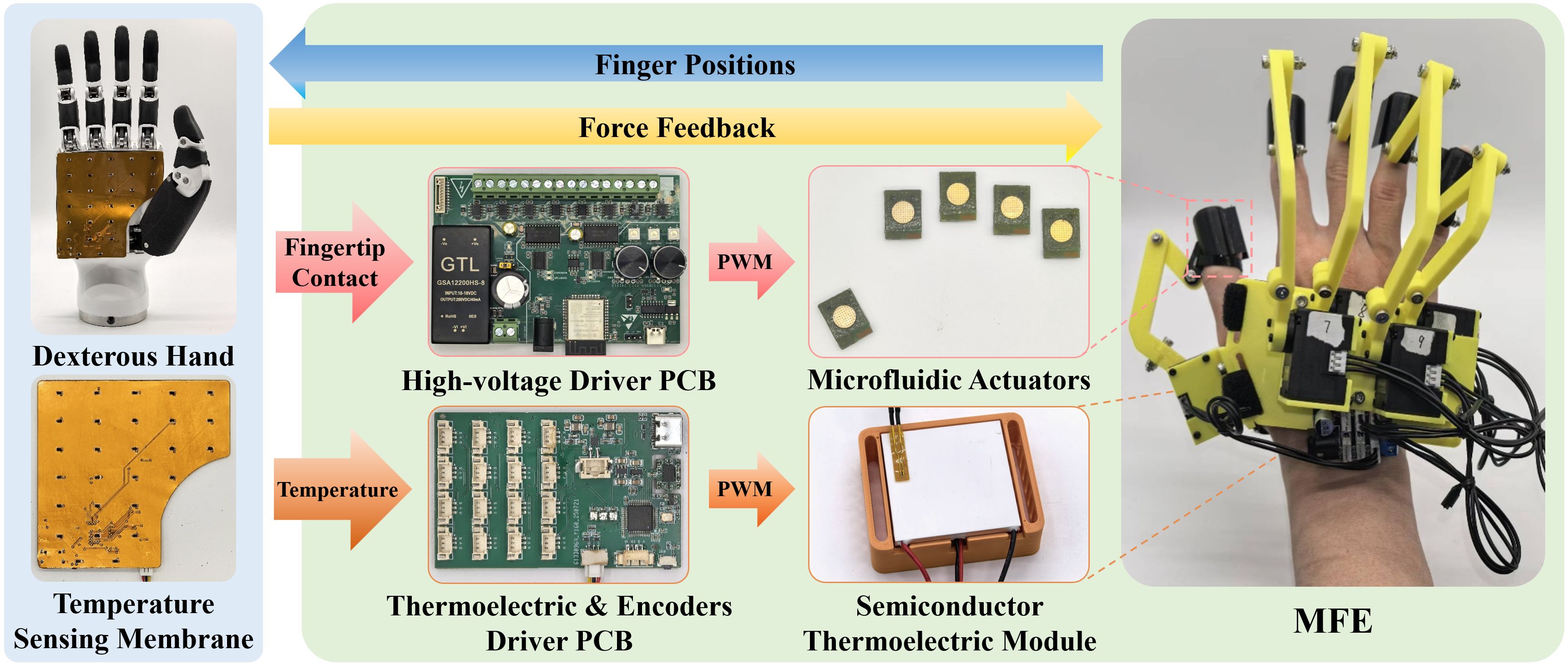}
    \caption{Technical pipeline for the MFE multimodal teleoperation system, encompassing the exoskeleton for finger's positional and force feedback, microfludic actuators for fingertip pressure rendering, thermoelectric module for temperature feedback.}
    \label{system}
    \vspace{-3mm}
\end{figure*}

This section describes the development of MFE, a hand exoskeleton designed to render multimodal haptic feedback. The methodology is presented in four thrusts: the design of the haptic exoskeleton, the fabrication of microfluidic actuators, the implementation of the temperature feedback, and integration into a bidirectional teleoperation system for haptic rendering. An overview of our system is shown in Fig.~\ref{system}.

\subsection{Haptic Hand Exoskeleton and Force Feedback}

\begin{figure}[h]
    \centering
    \includegraphics[width=0.88\linewidth]{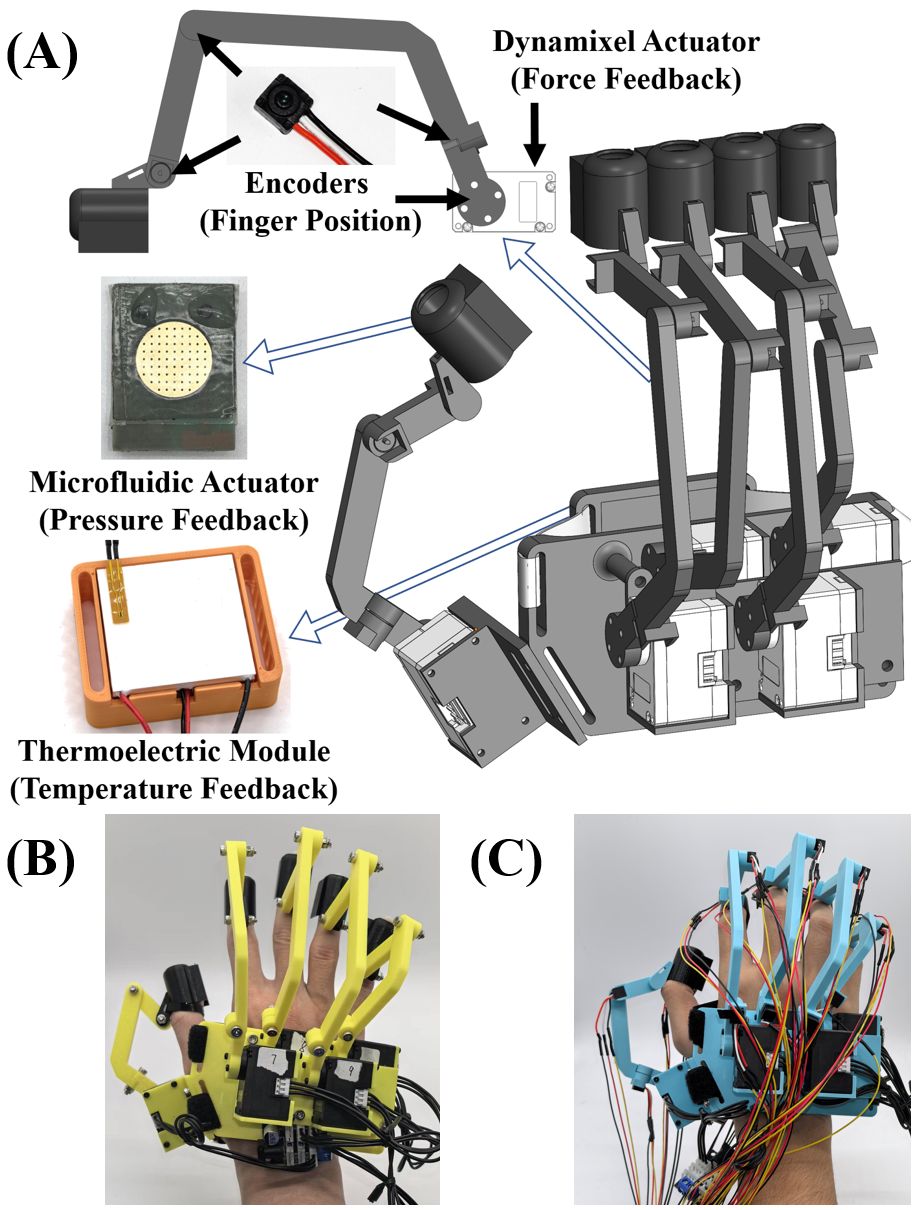}
    \caption{\textbf{MFE Exoskeleton:} (A) structures, (B) canonical design, and  (C) motion capture with additional joint encoders. }
    \label{fig:glove_structure}
\end{figure}

First, the mechanical structure of the haptic exoskeleton is designed to provide active force feedback to the fingers. This is accomplished using XL330-M288-T actuators (Dynamixel Inc.) in combination with 3D-printed linkage mechanisms. The actuators operate in torque control mode, enabling force feedback by specifying motor current, which allows them to apply the desired pushing or pulling force on the fingertips.

Efficient force transmission from the remote robot to the user is achieved through a linkage mechanism. Compared to recent works that use cable-driven systems \cite{lucidgloves, zhang2025doglove}, our design offers bidirectional controllable pushing and pulling forces, improved durability, reduced discomfort, and simplified assembly. Each finger's linkage mechanism consists of three flexion and extension joints and one lateral swing joint, allowing for straightforward forward-kinematics computation, intuitive motion, user comfort, and avoiding parallel configurations that may not be compatible with URDF description. The design of the finger mechanism is illustrated in Fig.~\ref{fig:glove_structure}(A). To accommodate users with different finger lengths, our open-source design includes multiple 3D-printable length options for ergonomic customization.

For fabrication, all linkage components and the base plate are 3D-printed using PLA. To enhance user comfort, the fingertip sections are printed with flexible TPU, allowing them to better conform to the user's fingertips. A slotted structure is included to integrate microfluidic actuators for rendering fingertip contact. The exoskeleton is secured to the user's hand using two Velcro straps, minimizing slippage and accommodating the installation of a thermoelectric module, which will be introduced in Sec.~\ref{sec:temp}. Sponge padding is attached to the contact surface of the exoskeleton and the back of the hand to enhance comfort. The entire assembly process is straightforward and can be completed in two hours.

The linkage mechanism also achieves finger motion capture, using both encoders inside Dynamixel actuators and additional in-place joint encoders. Encoders inside Dynamixel actuators alone provide 5 DoFs motion capture for fingers, which is sufficient to capture basic finger open/close motions  (Fig.~\ref{fig:glove_structure}(B)). For applications requiring exact finger positions, up to 15 additional encoders can be optionally installed at joints, resulting in the design shown in Fig.~\ref{fig:glove_structure}(C). These encoders are TOPKSA07W Hall sensors (TOPVR Inc.), capable of providing absolute angular measurements over a 360-degree range with 0.088° angular precision, enabling higher-fidelity tracking of joint rotations and enabling calculation of forward kinematics. To read the encoder's analog voltages, an STM32G473C8T6 microcontroller is used in conjunction with an XL4067 analog multiplexer. The connection between encoders and microcontroller is achieved by wires and MX1.25 terminal, considering the deformability required for connection. The total weight for canonical design is 315 grams, and the version with additional joint encoders is 348 grams (both include the weight of thermoelectric module).

\subsection{Microfluidic Actuators for Fingertip Feedback}

\begin{figure}[t]
    \centering
    \vspace{1mm}
    \includegraphics[width=0.95\linewidth]{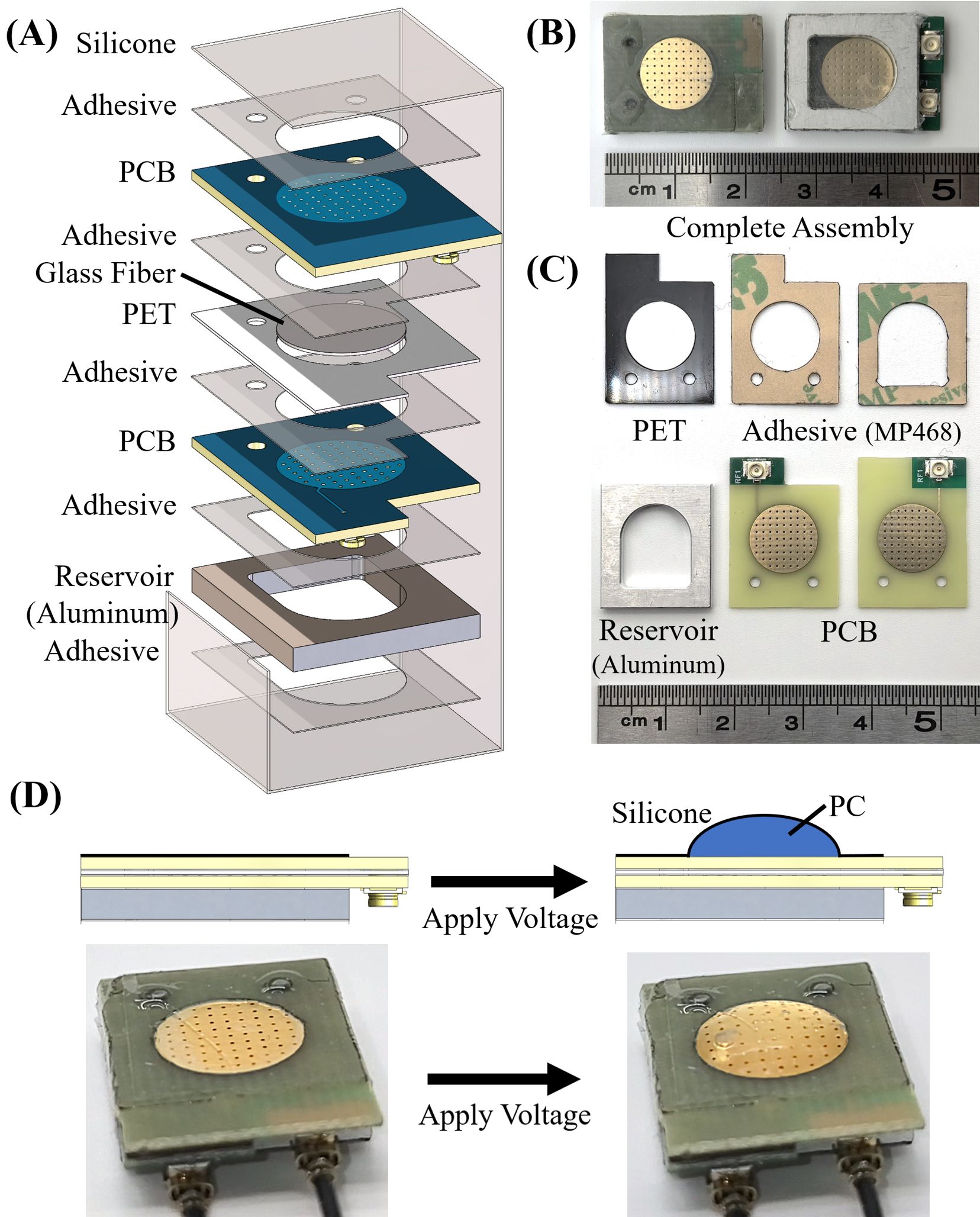}
    \caption{\textbf{Microfluidic Actuator.} (A) Exploded view, (B) Complete assembly, (C) Components, (D) Working principle.}
    \label{Microfluidic}
    \vspace{+1mm}
\end{figure}

To enhance MFE’s capability to render fingertip cutaneous and vibrotactile feedback, microfluidic actuators are fabricated and mounted onto MFE’s fingertip housings. We employ electroosmosis to generate controlled pressure and vibration, as electroosmotic actuators are uniquely capable of delivering both pushing and suction forces bidirectionally, without requiring bulky components such as motorized pumps. The fundamental design of this electroosmotic actuator is based on \cite{shultz2023flat, shen2023fluid}, which were originally developed to simulate fingertip contact in virtual reality environments. In our work, we adapt this approach for robot teleoperation and further optimize the pump design to increase the generated force.

Specifically, the device operates by establishing an electric field between two PCB boards, as illustrated in Fig.~\ref{Microfluidic}. This electric field induces fluid movement within microchannels, increasing internal pressure within a chamber. As a result, the membrane expands and applies pressure to the user's fingertip. When the direction of the electric field alternates periodically, the liquid flows can also cyclically along the field direction, producing vibrations perceptible at the fingertip. The device can generate vibrations with different frequencies and amplitudes via PWM-based voltage control, as detailed in Sec.~\ref{Sec:exp:microfluid} and supplementary video.

The fabrication process is shown in Fig.~\ref{Microfluidic}(A-C). The electrode components are two 0.8 mm thick PCBs, with fluid pathways implemented through vias of 0.2 mm diameter. Between the PCBs, a filter paper layer (Delvstlab Inc.) with 0.22 µm pore size was inserted to create the microchannels for the electro-osmotic effect. Compared to the 0.7 µm GF/F filter paper (Whatman Inc.) used in the original design \cite{shultz2023flat}, the finer pore size resulted in higher liquid flow rates and improved pumping performance. The layers are bonded together using 3M 468MP adhesive tape.

Propylene carbonate (PC) was selected as the working fluid due to its high dielectric constant and stability under strong electric fields, ensuring reliable and sustained device performance. During filling, air bubbles are removed from the microfluidic channels, as trapped air can interfere with fluid movement and degrade performance. To seal the device and create an elastic contact interface, PDMS silicone membranes were fabricated and attached to the top and bottom surfaces. Differ from the original design \cite{shultz2023flat}, a membrane thickness of 0.1 mm was chosen to provide suitable elasticity for large deformation while maintaining durability for liquid sealing. Also, a silicone surface treatment agent (J-750, JULE Inc.) was applied to bond the silicone membranes and the surfaces.

To control the actuators, we developed a driver PCB based on the HV513 high-voltage driver chip. The HV513 is powered by the GSA12200HS-8 power module (GTL-POWER Inc.), enabling operation at 200 V. High voltages generated by the driver board are delivered to the microfluidic actuators via IPEX-1 terminals and coaxial cables via bidirectional PWMs. For safety, a 3.3 k$\Omega$ current-sensing resistor is integrated into each channel to limit current, and monitor the current value. INA149 amplifiers were used to amplify current signals. 

\subsection{Temperature Sensing and Rendering} \label{sec:temp}

To facilitate the rendering of remote thermal sensations, the system integrates a temperature-sensing membrane and a thermoelectric heat pump module. The thermoelectric heat pump is suited for this purpose because it can generate both heat and cold. Building on the thermoelectric heat pump, we present a complete suite of haptic temperature rendering systems to promote such temperature rendering techniques.  Compared to previous work \cite{kim2020thermal}, our design targets an open-source system based on commercially available products and incorporates several advances, including closed-loop temperature control and an active cooling system to enhance heat transfer.

On the robot side, a temperature sensing membrane is designed to conform to the surface of the remote manipulator, as shown in Fig.~\ref{system} (left side). The circuit substrate is based on a flexible printed circuit (FPC), with an array of 27 miniature temperature sensors (NST1001-QDNR, NOVOSENSE Inc.), and shape fitted to the palm of Inspire’s RH56BFX robotic hand to enable full-palm temperature measurement. The sensed temperature is used for rendering on the user side and data collection for learning tasks. A STM32 microcontroller processes the sensor array data and transmits it to the computer via a USB virtual COM port.

On the user's side, the thermoelectric heat pump is attached to user's palm. To generate heat and cold, two PWM signals are sent via an H-bridge driver, which jointly offer a bidirectional voltage of  $\pm$5\,V. For closed-loop temperature control, an NTC thermistor is attached to the module surface. The NTC enables feedback control and constrains the heat pump's output temperature within a safe range of 10~°C to 55~°C. To improve temperature response time, a heat sink and cooling fan are mounted on the back of the thermoelectric module.

\begin{figure*}[h]
    \centering
    \includegraphics[width=0.92\linewidth]{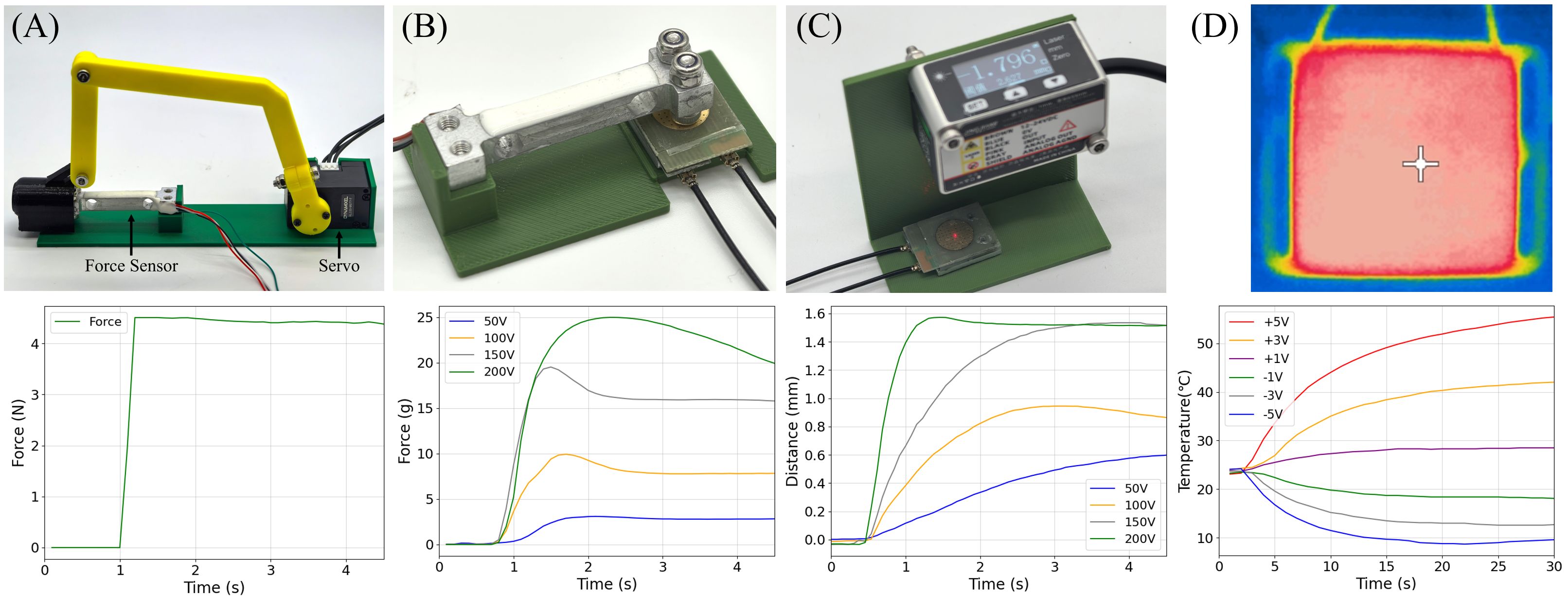}
    \caption{(A) Characterization of pulling force at finger's resting pose. Characterization of (B) maximum force, and (C) deformation distance generated by microfluid actuator (at 50, 100, 150, and 200 V, respectively). (D) Characterization of surface temperature of thermoelectric module (at +1, +3, +5, -1, -3, -5 V, respectively).}
    \label{experiment}
    \vspace{-5mm}
\end{figure*}

\subsection{Bidirectional Teleoperation Mapping System}
A robot teleoperation framework was developed to enable telepresence with multimodal haptic feedback. The system architecture is shown in Fig.~\ref{system}. The remote setup employs the RH56BFX robotic hand (Inspire Inc.), which has 6 controllable DoFs, one force sensor per finger, and a custom-developed temperature-sensing membrane on its palm. Each finger’s position is controlled by mapping encoder readings from MFE exoskeleton to the corresponding finger. Then, we render haptic information via the following approach. 

\textbf{Force Feedback in Bilateral Teleoperation:} Contact forces are delivered using a hybrid force–position control scheme. 
The robot hand is commanded to minimize the end-effector position error following \cite{qin2023anyteleop}:
\begin{equation}
\min_{\theta_{hand}} \| FK_{\text{hand}}(\theta_{hand}) - FK_{\text{MFE}}(\theta_{MFE}) \|,
\label{}
\end{equation}
where \( FK_{[\cdot]} \) denotes the forward kinematics, and \( \theta_{[\cdot]} \) represents the joint angles (on the MFE side, obtained from encoders).

On the robot side, when the contact force $F$ measured by the robot exceeds a predefined threshold $F_t$, 
the corresponding Dynamixel actuator in the MFE applies a reactive force to the user’s finger. 
The threshold $F_t$ is set to $1.47~\text{N}$ to suppress jitter and maintain control stability. 
The target current $I_{\text{target}}$ (in mA) for the Dynamixel actuator is computed as
\begin{equation}
I_{\text{target}} = k_I \, F,
\label{}
\end{equation}
where the force-to-current conversion factor $k_I$ is defined as
$k_I = \frac{1750~\text{mA}}{6000~\text{N}} \approx 0.292~\frac{\text{mA}}{\text{N}}$.

\textbf{Pressure Feedback:} When $F > F_t$, the microfluidic actuator is also activated to provide localized fingertip pressure feedback. The feedback intensity is determined by the following formula, where $D$ is the duty ratio of PWM signals and $k$ is a parameter for adjusting intensity. 
\begin{equation}
D = \mathrm{clip}\!\left( \frac{k}{1000} \,(F - F_t),\, 0,\, 1 \right).
\label{}
\end{equation}

\textbf{Thermo-haptic feedback:} 
During operation, the average temperature of the nine central sensors is used as the PID controller input for the thermoelectric module. The active cooling provided by the thermoelectric heat pump accelerates the glove’s surface temperature to stabilize within 3–6 seconds, ensuring rapid thermal rendering.

\section{Experiments}

Experiments were conducted to validate the effectiveness of our proposed MFE exoskeleton system. First, the performance of the force feedback mechanism, the microfluidic actuator and the temperature feedback framework was independently characterized. Next, the overall system was evaluated through human studies to demonstrate the usability of the device.

\subsection{Characterization of Force Feedback} \label{exp:force}

First, to comprehensively estimate pose-dependent force transmission and resistance, we analyzed the finger linkage using a kinematic model. The results show that under the servo’s maximum torque (0.52~Nm), the achievable output force across the workspace ranges from approximately 3.55 to 8.18~N. Similarly, considering the motor’s 0.03~Nm resistance torque, the peak back-driving resistance across the workspace is around 0.5~N. To further validate the estimated force, real qualitative experiments were conducted to evaluate the mechanism for rendering contact forces in the MFE. A force sensor was used to measure the maximum force at the end of the linkage, as illustrated in Fig.~\ref{experiment}(A). The linkage was positioned to match the finger's resting pose, and then the motor was actuated to its maximum stall torque capability. The results show that the actuator can provide forces up to 4.5~N at this pose. This value is consistent with theoretical analysis and well below typical human finger force limits to ensure safety (around 20~N~\cite{vergara2014introductory}).

\subsection{Characterization of Fingertip Pressure Feedback} \label{Sec:exp:microfluid}

Next, we characterized the response of the microfluidic actuator by assessing the generated force and the deformation distance of the surface. A force sensor was placed 0.1 mm above the microfluidic actuator. When the actuator was activated, the elastomer layer buckled, making contact with the force sensor. Then, the elastomer layer's deformation distance was measured using a laser distance meter (JINGJIAKE Inc.) positioned above the center of the silicone membrane.

The experimental setup and results are shown in Fig.~\ref{experiment} (B-C). The results revealed that both force and protrusion distance exhibited a step response pattern. At the maximum voltage of 200 V DC, the microfluidic actuator generated a peak force of 24 g (equivalent to 2.47 kPa), with some overshoot, and a maximum protrusion of 1.65 mm. Considering that human skin can distinguish pressure changes as low as 0.5 kPa \cite{Srinivasan1995Softness}, such pressure pulses are perceptible to the operator. Besides, our actuator can reach such 0.5 kPa threshold in approximately 0.1 s after activation. Although the actuator’s settling time is longer, user feedback in Sec.~\ref{userstudy} further confirms that the pressure variations and delay effects were not salient to the users and did not significantly affect the overall usability. We also quantified the sensor's response at different voltage levels. As the voltage decreased, both the generated force and displacement reduced, along with a corresponding decrease in overshoot, resulting in weaker haptic feedback. Therefore, we conclude that a higher voltage, such as 200 V DC, is more salient for effective haptic rendering.

\subsection{Characterization of Temperature Feedback}

We evaluated the effectiveness of thermoelectric heat pump in rendering remote temperatures. The thermoelectric module was driven at an averaged voltage levels of +1\,V, +3\,V, +5\,V, –1\,V, –3\,V, and –5\,V via bidirectional PWM, respectively. An infrared thermometer was positioned 10\,cm away from the center of the module to measure temperature changes.

As shown in Fig.~\ref{experiment}(D), when powered at +5\,V and –5\,V under ambient room temperature of 24~°C, the thermoelectric module achieved a steady-state temperature range between 10~°C and 55~°C. This range is both perceptually salient for thermal discrimination and safe for human contact.

\subsection{User Study Through Teleoperating a Robot Hand} \label{userstudy}

\begin{figure}[t]
\vspace{2mm}
    \centering
    \includegraphics[width=0.88\linewidth]{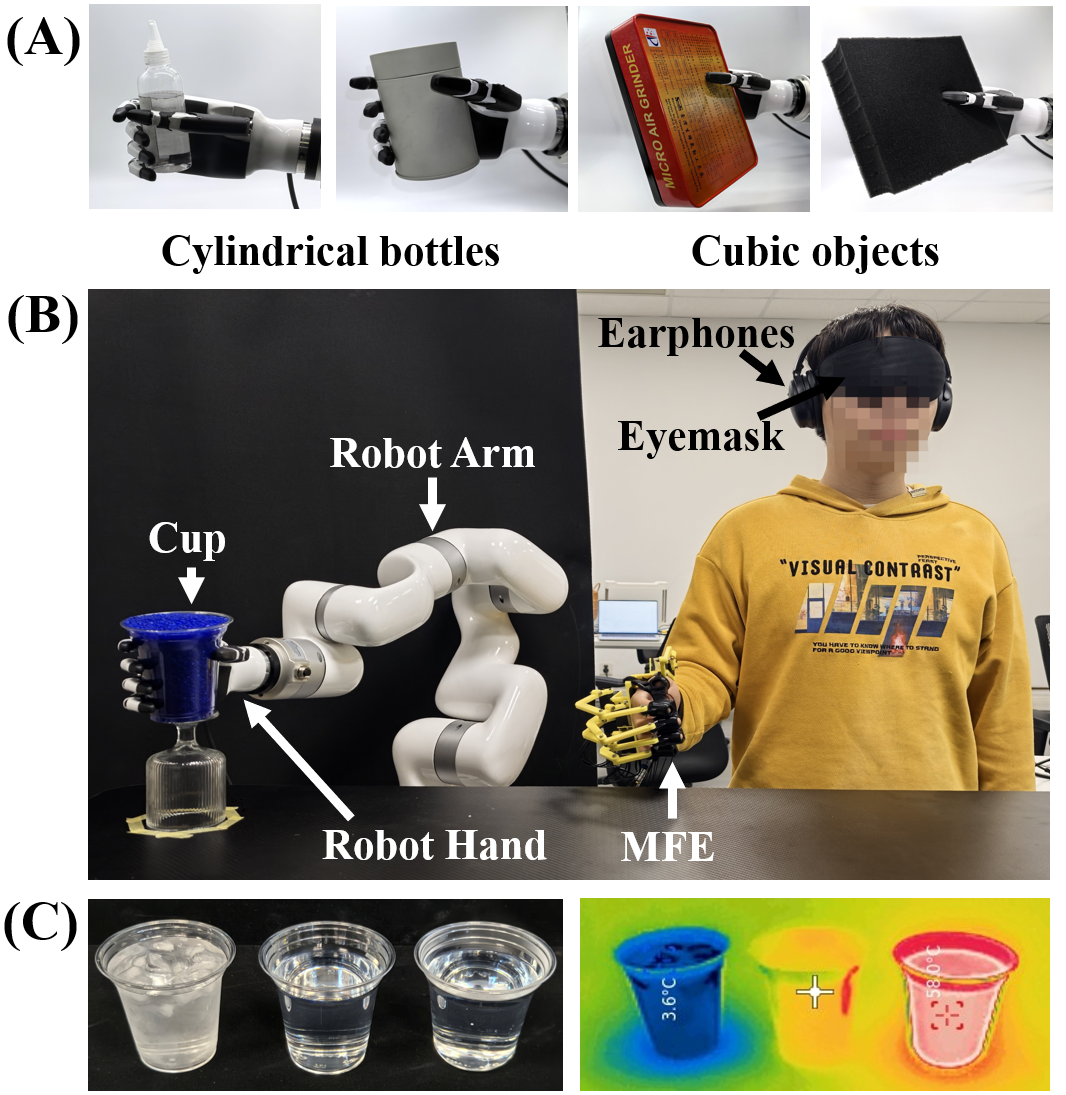}
    \caption{Experimental setup for (A) Task 1: distinguishing objects by shape and material stiffness, (B) Task 2: grasping a deformable cup filled with granular objects, and (C) Task 3: distinguishing temperatures of three cups of water.}
    \label{tasks}
    \vspace{+2mm}
\end{figure}

An user study was conducted to evaluate the usability of the MFE in contact-rich teleoperation tasks. Ten participants were asked to complete three tasks: (1) blindly distinguish between different objects using force feedback, (2) control the fingers of a robotic hand to hold a cup filled with granular material while the base robot arm was in motion, and (3) identify object temperatures through the MFE’s temperature rendering.

Participants relied solely on haptic feedback from MFE, with  vision and hearing blocked using blindfolds and noise-canceling headphones. All ten participants were between the ages of 18 and 25, had no physical disabilities, and performed the tasks using their dominant hand. The order of tasks was randomized to minimize potential learning effects.

\textbf{Task 1: Object Recognition.} Participants were asked to distinguish between two groups of objects: (1) Shape discrimination group: two rigid cylindrical bottles with different diameters, and (2) Stiffness discrimination group: two cubic objects (PEP material with Shore hardness 30 A, and tinplate box with Rockwell Hardness 50 B) with the same thickness but with varying stiffness, as shown in Fig.~\ref{tasks}(A). This task was performed under three sensory conditions: (1) force feedback only, (2) fingertip pressure feedback only, and (3) multimodal feedback (both force and fingertip feedbacks). 

The success counts for object differentiation are presented in Table~\ref{task1}. Using multimodal haptic feedback yielded the highest success rate in both tasks (100\% and 90\%, respectively), demonstrating its effectiveness. A decrease in success rate was observed when using force feedback (80\% and 60\%), and when using fingertip feedback alone (60\% and 40\%). These results suggest that multimodal feedback improves situational awareness during teleoperation that contributed to the task performance. We believe the difference in success rate mainly attributed to the discrepancy in output force range between actuators, since the motor can provide greater force and better constrain finger motion than microfluid actuators.

\begin{table}[h]
    \caption{Performance of 10 human subjects in object recognition task (Task 1). Numbers are successful trails. } 
    \centering
    \begin{tabular}{c | c c c}
        \hline\hline
          & Force & Fingertip & Multimodal \\
        \midrule
        Obj. Shape  & 8/10 & 6/10  & \textbf{10/10} \\
        Obj. Stiffness & 6/10 & 4/10  & \textbf{9/10} \\
        \hline\hline
    \end{tabular}
    \label{task1}
    \vspace{+1mm}
\end{table}

\textbf{Task 2: Deformable Object Manipulation.} We further evaluated the force and pressure feedback through a deformable object manipulation task. Participants were asked to steadily grasp a deformable plastic cup filled with silicone gel granules. The robot arm move in a preprogrammed procedure and create inclination up to 10 degrees in various directions. As the robot arm moved, participants had to dynamically adjust their gripping force to prevent the granules from spilling, as shown in Fig.~\ref{tasks}(B). The challenge was to minimize deformation of the cup to avoid spills, while maintaining a secure grip to prevent the cup from falling.

\begin{figure}[ht]
    \centering
    \includegraphics[width=0.82\linewidth]{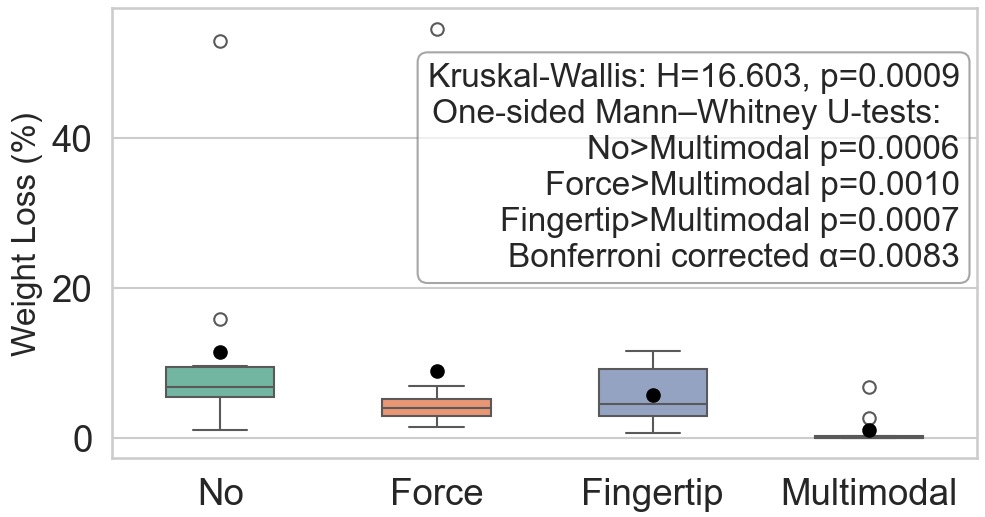}
    \caption{Performance of 10 human subjects in manipulating cups with granular objects (Task 2), and statistical test results.}
    \label{fig:task2}
\end{figure}

To facilitate comparison, the task was evaluated under four conditions: no feedback, force feedback, fingertip feedback, and multimodal feedback. Weight loss due to spilled silicone granules is shown in Fig.~\ref{fig:task2}. Because the data exhibit non-uniform distributions and outliers, we employed a non-parametric Kruskal–Wallis test to assess overall group differences ($p<0.05$). One-sided post-hoc Mann–Whitney U-tests with Bonferroni correction were then used to test our directional hypothesis that multimodal feedback performs better. The results show that multimodal feedback yields significantly lower weight loss than all other conditions (Fig.~\ref{fig:task2}), while differences among the remaining conditions are not significant. Cup drops occurred only with no feedback (two trials) or with force feedback (one trial), and none with multimodal feedback. This demonstrates improved teleoperation performance.

The benefits of each haptic modality were assessed through questionnaires. In both tasks, all 10 participants reported feeling more confident in completing the task with multimodal feedback. All 10 subjects found the force feedback mechanism helpful, while 8 out of 10 subjects also found the fingertip feedback beneficial. These results highlight the effectiveness of multimodal haptic feedback in handling deformable objects.

\textbf{Task 3: Temperature Recognition.} The effectiveness of our temperature feedback framework was evaluated through two temperature recognition experiments. First, participants were asked to remotely rank three cups of water in order of temperature from highest to lowest. The temperatures of the cups were set to 4~°C, 20~°C, and 60~°C, as illustrated in Fig.~\ref{tasks}(C). Second, participants were asked to report the moment they perceived a temperature change. In this task, two cups of water at 4~°C and 60~°C were sequentially brought into contact with the temperature-sensing membrane. The goal was to measure the time delay between the sensor detecting a temperature change and the participant perceiving it.

In our experiment, all participants successfully sorted the water cups according to the rendered temperature, confirming the usability. On average, participants required 4.15 s to perceive an increase in temperature and 3.88 s to perceive a decrease. This perceptual delay corresponds to a distinguishable temperature difference of approximately 5~°C.

\subsection{Limitations and Future Work}

 Limitations were observed during experiments. One limitation is the delay of temperature haptic rendering: users require around 4 seconds to distinguish a remote temperature differences. Our investigation suggests this delay is mainly due to the specific heat capacity of the thermoelectric heat pump and the limited heating/cooling power in the current design. This limitation could be mitigated by using a smaller size heat pump device and a higher supply voltage. Second, motor and microfluidic actuator also has a response time in requires hundreds of milliseconds to reach its peak response. Such latency may limit usability in high-speed teleoperation tasks. Third, future work could focus on integrating stretchable FPCs for adopting a bus-based communication protocol to reduce wiring complexity with encoders. Last, since only a single thermoelectric module is used, the system currently lacks spatial resolution for temperature rendering over larger contact areas. Future work will focus on developing thermoelectric arrays to achieve high-resolution temperature feedback. 

Overall, our MFE system is the first open-source platform capable of concurrently rendering force, pressure, and temperature. Experiments demonstrate its ability to enhance user situational awareness across these tactile modalities. Given MFE’s potential to improve data collection quality in teleoperation, we plan to leverage it for data gathering to advance embodied AI research in future studies.

\section{Conclusions}
To achieve high-fidelity haptic feedback, this paper presented MFE, an exoskeleton system designed to deliver comprehensive force, pressure, and temperature feedback to users' hands. The system employs force-controlled actuators for rendering contact force, and microfluidic flat actuators for rendering skin deformation via constant pressure or vibration. The proposed hardware can generate 3.5-8.1~N of force per finger and exert a maximum pressure of 2.47 kPa to deform the fingertip skin. Also, our system offers temperature feedback on the palm to represent temperature information, achieving temperature rendering from 10~°C to 55~°C.

A teleoperation system was developed to evaluate the proposed exoskeleton design. Experimental results demonstrated that the system effectively provides multimodal haptic feedback. With the MFE, participants achieved higher success rates in object recognition and deformable object grasping tasks. Questionnaires revealed increased confidence in task completion when haptic feedback was provided. Future work will focus on delivering more accurate and realistic feedback, simplify cable connection, and exploring MFE for assisting multimodal data collection for robot imitation learning.

\section*{Acknowledgment}
We thank Yutao Ming for exploring a haptic glove for virtual reality \cite{lucidgloves} and Xi Zhang for assistance with the photo shoot.

\bibliography{egbib}
\bibliographystyle{IEEEtran}
\end{document}